\documentclass[letterpaper, 10 pt, conference]{ieeeconf}  

\IEEEoverridecommandlockouts                              

\title{\LARGE \bf
Polar Collision Grids: Effective Interaction Modelling for Pedestrian Trajectory Prediction in Shared Space Using Collision Checks*}

\author{Mahsa Golchoubian$^{1\dag}$, Moojan Ghafurian$^{2}$, Kerstin Dautenhahn$^{2}$, Nasser Lashgarian Azad$^{1}$
\thanks{* This research was undertaken, in part, thanks to funding from the Canada 150 Research Chairs Program and the Natural Sciences and Engineering Research Council of Canada (NSERC).}
\thanks{\dag Corresponding Author: Mahsa Golchoubian {\tt\small mahsa.golchoubian@uwaterloo.ca}}
\thanks{$^{1}$Department of Systems Design Engineering, University of Waterloo, Canada}
\thanks{$^{2}$Department of Electrical and Computer Engineering, University of Waterloo, Canada}%
}

\usepackage{graphicx}
\usepackage{subfigure}
\usepackage{tablefootnote}
\usepackage{dblfloatfix}  
\usepackage{amsmath}
\usepackage{dsfont}
\usepackage{breqn}
\usepackage{float}

 \usepackage[pscoord]{eso-pic}
\newcommand{\placetextbox}[3]{
 \setbox0=\hbox{#3}
 \AddToShipoutPictureFG*{ \put(\LenToUnit{#1\paperwidth},\LenToUnit{#2\paperheight}){\vtop{{\null}\makebox[0pt][c]{#3}}}
 }
 }

  \placetextbox{.5}{0.055}{\footnotesize{© 2023 IEEE. Personal use of this material is permitted. Permission from IEEE must be obtained for all other uses, in any current or future media,}} 
  \placetextbox{.5}{0.045}{\footnotesize{including reprinting/republishing this material for advertising or promotional purposes, creating new collective works, for resale or redistribution}}
  \placetextbox{.5}{0.035}{\footnotesize{to servers or lists, or reuse of any copyrighted component of this work in other works.}}

\begin{document}

\maketitle
\thispagestyle{empty}
\pagestyle{empty}

\begin{abstract}

Predicting pedestrians' trajectories is a crucial capability for autonomous vehicles' safe navigation, especially in spaces shared with pedestrians. Pedestrian motion in shared spaces is influenced by both the presence of vehicles and other pedestrians. Therefore, effectively modelling both pedestrian-pedestrian and pedestrian-vehicle interactions can increase the accuracy of the pedestrian trajectory prediction models. Despite the huge literature on ways to encode the effect of interacting agents on a pedestrian's predicted trajectory using deep-learning models, limited effort has been put into the effective selection of interacting agents. In the majority of cases, the interaction features used are mainly based on relative distances while paying less attention to the effect of the velocity and approaching direction in the interaction formulation. In this paper, we propose a heuristic-based process of selecting the interacting agents based on collision risk calculation. Focusing on interactions of potentially colliding agents with a target pedestrian, we propose the use of time-to-collision and the approach direction angle of two agents for encoding the interaction effect. This is done by introducing a novel polar collision grid map. Our results have shown predicted trajectories closer to the ground truth compared to existing methods (used as a baseline) on the HBS dataset.

\end{abstract}

\section{Introduction}

Autonomous vehicles (AVs) are envisioned to be part of our everyday future transportation system. For an efficient and safe operation of these AVs in spaces where pedestrians are present, they should be able to predict the future trajectories of nearby pedestrians.

Accurate pedestrian trajectory prediction requires the integration of different sources of information. This includes the past trajectory history of each agent as well as the between-agent interaction effects (e.g., how one pedestrian's movement may affect another pedestrian's). Due to the complexity of pedestrians' motion in interactive settings, many deep-learning models have been proposed in the literature for predicting pedestrian trajectories while considering the interaction effects (e.g.,\cite{alahi2016social,gupta2018social,xue2018ss,salzmann2020trajectron++,lee2017desire,cheng2020mcenet}). Most of these methods have focused their training on datasets that only include pedestrians without any vehicle present \cite{alahi2016social,gupta2018social,xue2018ss}. Thus, these past models have only accounted for pedestrian-pedestrian interaction effects. However, such models cannot directly be used for predicting a pedestrian's trajectory in the presence of pedestrians and vehicles.

Recent work has started considering the effect of pedestrian-vehicle interaction in the prediction models as a distinct influencing factor \cite{zhang2022learning,eiffert2020probabilistic,cheng2018modeling,hu2020collaborative,mo2022multi,li2021spatio,salzmann2020trajectron++}.
Examples are using two separate pooling modules for information aggregation of pedestrians and vehicles that are interacting with the target pedestrian \cite{zhang2022learning}, or are accounting for different edge types in a graph neural network (GNN) architecture for distinguishing the two types of interaction apart \cite{mo2022multi,salzmann2020trajectron++}.
A complete review of these pedestrian trajectory prediction models that also account for pedestrian-vehicle interaction effects is provided in \cite{10181234}. 

In deep learning approaches for modelling interaction effects, two main questions need to be answered: (1) How interacting agents are specified, and
(2) How to formulate the effect of these interacting agents on the pedestrians' future trajectories.
Most of the existing literature in this area has focused only on proposing different methods for answering the second question. Simplified answers have been given to the first question such as considering agents as interacting agents that are closer to each other than a specified threshold  \cite{alahi2016social,bi2019joint,cheng2020mcenet,lee2017desire,carrasco2021scout,li2021spatio,girase2021loki}, or considering all the agents in the scene and aggregating all their trajectory information for formulating interaction \cite{eiffert2020probabilistic,zhang2022learning,hu2020collaborative}.
While some proposed methods have considered learning the importance of each neighbouring agent through weightings learned in attention layers \cite{eiffert2020probabilistic, lai2020trajectory}, still, a measure for selecting the interacting agents in the first place is missing in the literature. For example, in classical heuristic-based methods, pedestrians' motions when encountering vehicles are designed to be governed by time-dependent collision risk features such as time to collision \cite{predhumeau2022agent,predhumeau2021agent} or minimum gap acceptance \cite{cheng2019study,li2015studies} (defined as the time left between the pedestrian and the nearest upcoming vehicle).

Getting inspired by these methods, in this paper, we are proposing the use of Time To Collision (TTC) for specifying interacting agents (both pedestrians and vehicles) that can influence a target pedestrian's future trajectory. 

Moreover, in response to the second question about formulating the possible effect of the interacting agents on a target pedestrian's trajectory, most of the existing models rely on relative distance features such as pooling the information of all interacting agents in a fixed grid map around the target pedestrian \cite{alahi2016social,bi2019joint,xue2018ss,cheng2020mcenet,lee2017desire}, or aggregating the information of the neighbouring agents using their relative distance to the target pedestrian \cite{zhang2022learning, gupta2018social,zuo2021map,lai2020trajectory, eiffert2020probabilistic}. In all these distance-dependent methods of capturing interaction, the effects of approach directions and speed are missing while these two factors can indeed have a high impact on pedestrians' motion behaviour \cite{huber2014adjustments,vemula2018social}. 
In other words, two agents could be close, in terms of relative distance, but are not necessarily influencing each other's trajectory, if walking in parallel or getting away from each other.

Few other models \cite{lai2020trajectory,li2021spatio,mo2022multi} have considered the raw velocity information as an additional agent state. However, velocity information can be used for deriving more explainable engineered interaction features.
Therefore, in this paper, we will take advantage of the information embedded in the velocity vectors to calculate the collision probabilities, and define a polar collision grid map for capturing interaction features that can influence a pedestrian's future trajectory.

We believe that a neural network model can learn interaction-influenced motion patterns more effectively if given more explainable engineered features as input. 

We can summarize our contribution as follows:

\begin{enumerate}
    \item Specifying interacting agents based on TTC as a more effective indicator compared to the relative distance used in the literature for studying interaction effect in pedestrian trajectory prediction 
    \item Introducing a novel polar collision grid map for capturing interacting effects in a deep-learning-based prediction model using both TTC and approaching angle information.
\end{enumerate}

\section{Related Work}

In deep learning-based pedestrian trajectory prediction models, an interaction encoder is responsible for building a feature that captures the between-agent interaction effects on the future trajectory. Various structures have been proposed for this interaction encoder in the literature which is reviewed here from two main perspectives: 1) The selection of interacting agents, 2) The encoded interaction features.

\subsection{Specifying Interacting Agents}

The deep learning methods used in the literature often rely on positional distance for reasoning about the interacting agents. In \cite{alahi2016social,bi2019joint,cheng2020mcenet,lee2017desire} an occupancy grid map centred at each agent's position is used for constructing the interaction feature of that agent. With a fixed grid size, only agents closer than a threshold to that target agent are considered as having an interaction and are therefore considered for constructing that target agent's interaction feature, thus, only focusing on the effects caused by close-by agents. 

In the GNN models proposed in \cite{carrasco2021scout, li2021spatio, girase2021loki}, an agent's node is connected only to other nodes within a certain distance and the interaction effects of only those connected nodes are considered in the message passing process.

To overcome the limitation of these methods in capturing possible farther-away interactions, others have included the information of all agents in the scene for building the interaction feature \cite{eiffert2020probabilistic,zhang2022learning},  or considered a fully connected graph in the GNN models used for trajectory prediction \cite{hu2020collaborative}.

In a novel approach proposed by Li et al. (2021), the  existence of an interaction edge in a GNN model is decided through a reinforcement learning (RL) framework on top of the prediction model \cite{li2021rain}. In their RL formulation, the action is specified as making the edge between the two nodes on or off and the reward is defined based on how close the prediction of the selected graph connection is to the ground truth trajectory.

Su et al. (2022), have proposed different connectivity graphs for interaction consideration in their GNN structure namely \textit{View graph}, \textit{Direction graph} and \textit{Rate graph} \cite{su2022trajectory}. In the \textit{View graph} only agents present in a target pedestrian's field of view are connected to that pedestrian's node in the graph and are considered to have an interaction effect on the pedestrians. For \textit{Direction graph} and \textit{Rate graph}, these edges are decided based on whether the two agents' moving direction has a crossover \cite{su2022trajectory}. 
While the \textit{Direction graph} proposed in \cite{su2022trajectory} can potentially account for the collision risks, it should be pointed out that not every path crossover necessarily yields a collision. The agents on two crossing paths will not collide if the two agents get to that crossover point at different times. Therefore, we propose the direct use of collision checks for specifying the interacting agents while also accounting for the time alignments in the trajectories for defining a collision.

\subsection{Interaction Features}

Regardless of how the potentially interacting agents are selected, the interaction features used throughout the literature are again mainly constructed based on the relative distances of the neighbours to the target agent.

In \cite{alahi2016social,bi2019joint,xue2018ss,cheng2020mcenet,lee2017desire}, fixed-size grid maps centred at the target agents' locations is defined and an interaction tensor is built using the occupied cell of each neighbour according to relative distances to the target agent. Both rectangular shape grids \cite{alahi2016social,bi2019joint,xue2018ss} and circular shape (polar) grids \cite{cheng2020mcenet,lee2017desire,xue2018ss} have been proposed. But even in the polar grids proposed in \cite{cheng2020mcenet,lee2017desire,xue2018ss}, only the angle of the distance vector between the two agents is accounted for in the grid construction without caring about the approach direction of the two agents.

In other models the relative distances between the target agent and its neighbours are embedded and concatenated with the hidden state of the neighbours and finally, this concatenated feature from all the neighbours is aggregated using a softmax \cite{eiffert2020probabilistic} or a max pooling layer \cite{zhang2022learning, gupta2018social,zuo2021map} to construct the interaction feature. In the soft attention mechanism used in \cite{lai2020trajectory} for aggregating the neighbours' hidden states, the bearing angle between the two agents is also used along with their distance in the calculation of the attention score for each neighbour.

Cheng et al. (2018) implemented a novel approach of using collision probability between each target agent and all its neighbours for building the interaction feature tensor inputted to an LSTM model \cite{cheng2018modeling}. But again their collision probability is calculated as an exponential function of the relative distance, ignoring the influence of speed. 

Therefore, in our work, we are proposing a collision probability grid that stores actual TTC, and approaching angle information by using both velocity and position data.

\section{Method}

\subsection{Problem Definition}

Pedestrian trajectory prediction is defined as predicting the future positions of pedestrians given a short observed history of their own and the close-by agents' (pedestrians and vehicles) trajectories \cite{alahi2016social}. The position of pedestrians and vehicles at time step $t$ along their trajectories are denoted as $X_t^{ped} = [X_t^1, X_t^2, ..., X_t^{n_p}]$ and $X_t^{veh} = [X_t^1, X_t^2, ..., X_t^{n_v}]$, where ``ped" stands for pedestrian and ``veh" stands for vehicle, and $n_p$ and $n_v$ are the numbers of pedestrians and vehicles in the scene respectively. The x-y position coordinate of pedestrian or vehicle $i$ at time $t$ in this formulation is defined as $X_t^i = (x_t^i,y_t^i)$. The velocity vector of each agent denoted as $V_t^i = ({v_x}_t^i,{v_y}_t^i)$
can also be calculated using the position difference between successive time steps, knowing the timestamp between the captured frames.

Given the observed trajectories of all agents for the time steps between $0<t\leq T_{obs}$, we predict the trajectory of the pedestrians across the future time period of $T_{obs} < t\leq T_{pred}$ as $\tilde{Y}_t = [\tilde{Y}_t^1, \tilde{Y}_t^2, ..., \tilde{Y}_t^{n_p}]$. The ground truth trajectory of pedestrians for any time step in this prediction period is denoted as $Y_t = [Y_t^1, Y_t^2, ..., Y_t^{n_p}]$. The predicted position coordinates $\tilde{Y}_t^i$ are derived from the predicted Gaussian distribution output of the model $\hat{Y}_t^i$, which will be described later in Eq. \ref{eq:Normal}.

\subsection{Overall Framework of the Trajectory Prediction Model} 

\begin{figure}
  \centering
  \includegraphics[width=0.8\linewidth]{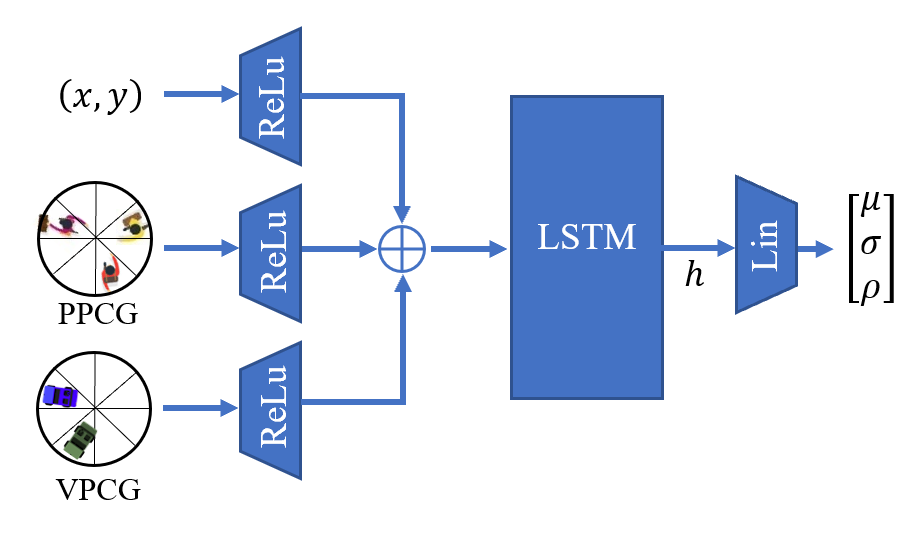}
  \caption{The overall framework of our proposed pedestrian trajectory prediction. The embedded spatial feature, the pedestrian-pedestrian interaction and the pedestrian-vehicle interaction features are concatenated before being inputted into the LSTM module.}
  \label{fig:Overall}
\end{figure}

Our overall approach for predicting pedestrians' trajectories in an environment shared with both pedestrians and vehicles is shown in Fig. \ref{fig:Overall}. The movement pattern in the pedestrian's trajectories is learned through an LSTM Network. The network receives three types of input features for each pedestrian: (1) the spatial embedded feature of the pedestrian, (2) the embedded pedestrian-pedestrian interaction feature, and (3) the embedded pedestrian-vehicle interaction feature. These features are aggregated together and used as the input to the LSTM at each timestep.

Similar to other works \cite{alahi2016social,eiffert2020probabilistic}, instead of predicting the actual future positions, our model will predict the parameters of a bi-variant Gaussian distribution of possible future positions to account for prediction uncertainty. Therefore, the output of our LSTM goes through a linear layer to predict the means ($ \mu_x, \mu_y$), standard deviations ($ \sigma_x, \sigma_y$) and the correlation coefficient ($\rho$) of the bi-variant Gaussian distribution.

\subsubsection{Spatial embedded feature}

The effect of each pedestrian's own trajectory history on the predicted next position is captured through this feature where the spatial coordinate of the pedestrian is embedded through a single layer with ReLU non-linearity. In the following equation, this layer is denoted as $\phi_e(.)$ with $W_e$ being its embedding weights.

\begin{equation}
\label{eq:Spatial} 
    e_t^i = \phi_e(\Delta X_t^i; W_e)
\end{equation}

The spatial input will be provided as displacements $\Delta X_t^i = (x_t^i - x_{t-1}^i, y_t^i - y_{t-1}^i)$ which is the relative distance between pedestrian $i$'s current time step $t$ and the previous time step $t-1$.

\subsubsection{Interaction embedded features}

How interactions with other agents (i.e., pedestrians and vehicles) can influence a pedestrian's future trajectory is captured through the interaction feature input. 
These interactions are captured through a polar collision grid map in form of a tensor that will be discussed in more detail in Section \ref{InterFeat}. We have separate tensors for encoding pedestrian-pedestrian (PPCG) and pedestrian-vehicle interaction (VPCG), as a vehicle can influence a pedestrian's trajectory differently from how another neighbouring pedestrian does. Each of these two tensors will be processed by going through an embedding layer with ReLU nonlinearity ($\phi_p$ for ped-ped and $\phi_m$ for ped-veh\footnote{Ped stands for pedestrians. Veh stands for vehicle} interaction) before being used as an input to the LSTM network. 

\begin{equation} \label{eq:PedInt} 
    p_t^i = \phi_p(PPCG_t^i; W_p)
\end{equation}
\begin{equation} \label{eq:VehInt} 
    m_t^i = \phi_m(VPCG_t^i; W_m)
\end{equation}

\subsubsection{Prediction module}

The three inputs of spatial feature ($e_t^i$), ped-ped interaction feature ($p_t^i$) and ped-veh interaction feature ($m_t^i$) will be concatenated and used as input to the LSTM cell with $W_l$ weights.

\begin{equation} \label{eq:lstm} 
    h_t^i = LSTM(h_{t-1}^i, concat(e_t^i, p_t^i, m_t^i); W_l)
\end{equation}

The output of the LSTM will then go through a linear layer with $W_o$ embedding weight to output the parameters of the bi-variant Gaussian distribution. The predicted position can then be generated by sampling from this distribution. 

\begin{equation} \label{eq:Outlay} 
    \hat{Y}_t^i = [\mu_x^i, \mu_y^i, \sigma_x^i, \sigma_y^i, \rho^i]_t = W_o h_t^i
\end{equation}
\begin{equation} \label{eq:Normal} 
    \tilde{Y}_{t+1}^i \sim \mathcal{N}(\hat{Y}_t^i)
\end{equation}

\subsubsection{Loss function}

Given that the model is predicting the parameters of a bi-variant Gaussian distribution ($\hat{Y}_t^i$) over the next positions, a negative log-likelihood loss function is used for learning the parameters of the model as follows:

\begin{equation} \label{eq:loss} 
    Loss = - \sum_{i,t} log(P(Y_{t+1}^i | \tilde{Y}_{t+1}^i))
\end{equation}

\subsection{Interacting Agents}  \label{InterAgent}

In this paper, we rely on collision risks for indicating each target pedestrian's nearby interacting agents. Only these specified interacting agents will then be considered for constructing the interaction feature of each target pedestrian. This method is opposed to other approaches followed in the literature that consider either all the agents present in the scene or rely only on distance for specifying interacting agents and building the interaction feature. The intuition is that at each time step a pedestrian's trajectory will be affected only by those agents that are in a collision course with it if the pedestrian keeps its current trajectory trend. 

We use the feature of Time To Collision (TTC) as a proxy for measuring the collision risk with others. Therefore, at each time step, having the position and velocity vectors we calculate the time left for the agents to get closer to each other than a minimum comfortable distance ($d_{min}$), which we define as collision. This time between agent $i$ and $j$ with position vectors $(x, y)$ and velocity vectors $(v_x, v_y)$ is calculated according to eq \ref{eq:tconf}. The notations used are for the information at time step t but we have removed the subscript $t$ for simplicity.

\begin{dmath}\label{eq:tconf}
    TTC^{ij} = \\ \frac{- (\vec{D}_{rel} \cdot \vec{V}_{rel}) \pm  \sqrt{(\vec{D}_{rel} \cdot \vec{V}_{rel})^2 - |\vec{V}_{rel}|^2 (|\vec{D}_{rel}|^2 - d_{min}^2)}}{|\vec{V}_{rel}|^2}
\end{dmath}

In the above equation, the vectors for relative distance and the relative velocity between agents $i$ and $j$ are denoted as $\vec{D}_{rel} = X^i - X^j$ and $\vec{V}_{rel} = V^i - V^j$ respectively. Moreover, (.) denotes the dot product of two vectors and $|\vec{D}|$ is the size of the vector $\vec{D}$. 

Using equation \ref{eq:tconf}, all the agents with a valid TTC of greater than zero and less than a specified threshold of $TTC_{thre}$ will be considered for constructing the interaction features as discussed in the next section. For any agent that might already be within $d_{min}$ distance of the target pedestrian a TTC of zero will be considered.

\subsection{Polar Collision Grid Interaction Feature}  \label{InterFeat}

At each time step, having the specified interacting agent and their TTC, we construct the interaction feature in form of a polar grid centred at each target pedestrian's current position and in the direction of its current velocity vector. 

We follow a similar approach to how we relied on collision risk for specifying interacting agents for building the interaction effect feature. 

With the justification that a pedestrian follows relatively similar evasive manoeuvres for agents in the same collision course in terms of approaching angle, we consider the angle between the velocity vectors of two potentially colliding agents as another feature for encoding interaction and learning the trend of the evasive behaviour. Therefore, we discretize the angular space around each pedestrian's current velocity vector to a specified number of sectors ($n_{sector}$) and for each conflicting agent, we calculate the discretized sector where its approach angle lies on. The angles are calculated with respect to the target pedestrian's current velocity vector in a counter-clockwise direction. This process results in a polar grid map around each pedestrian's current velocity vector while specifying the presence of each interacting agent in each cell of this polar grid.

To convert this polar grid map into an interaction feature that also includes collision risk information, we store the $TTC_{thr}-TTC$ value of each interacting agent in its corresponding cell. We aggregate the information of the multiple interacting agents by focusing on the riskiest agent in each approach angle cell which is the agent with the lowest time to collision or the highest value of $TTC_{thr}-TTC$. The resulting grid will be called the Polar time-to-Collision Grid (PPCG) which is a tensor of size $1 \times n_{sector} $ calculated according to equation \ref{eq:PCG} for each pedestrian $i$ at time step $t$.  In this equation, $\mathds{1}_n[\theta]$ is a binary indicator function to specify if $\theta$ (the angle between vector $\vec{V}^i_t$ and $\vec{V}^j_t$) is in the $n^{\text{th}}$ cell of the grid, and $N^i$ is the set of all interacting agent of pedestrian~$i$.

\begin{equation} \label{eq:PCG} 
    PPCG^i_t(n) = \max_{j \in N^i} (\mathds{1}_n[\angle (\vec{V}^i_t, \vec{V}^j_t)] (TTC_{thr} - TTC^{ij}))
\end{equation}

\begin{figure}
  \centering
  \includegraphics[width=1.0\linewidth]{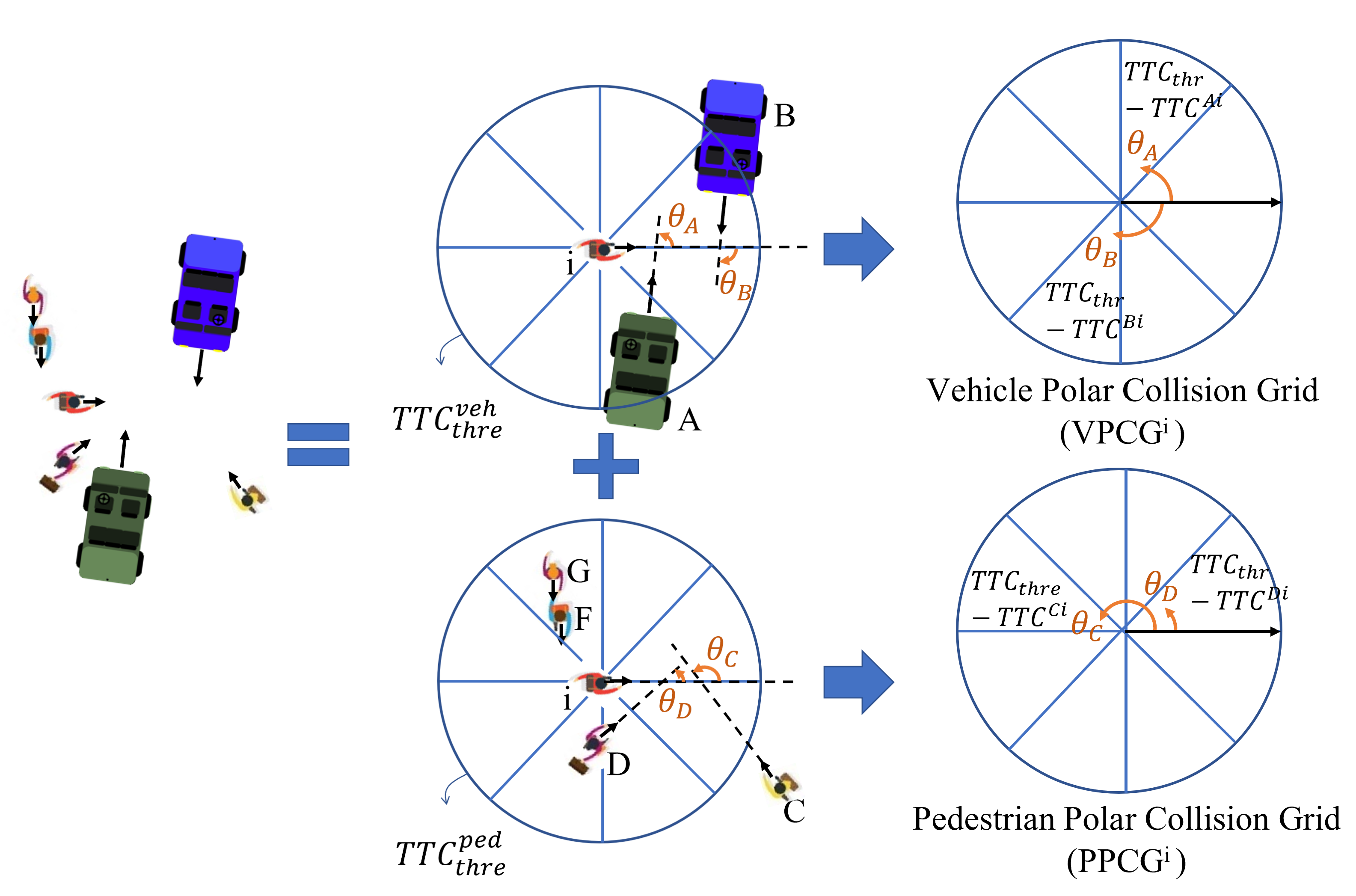}
  \caption{The process of building the interaction features for pedestrian $i$ according to the neighbouring agent's time to collision (TTC) and approach direction. Eight discrete sectors around the target pedestrian are considered in this figure for the approach angles. Only neighbouring agents with a TTC lower than a threshold will be considered in the construction of the interaction feature. The interaction feature for neighbouring vehicles and pedestrians are constructed separately.}
  \label{fig:InteFea}
\end{figure}

This tensor will be constructed separately for pedestrians (PPCG) and vehicles (VPCG) that are interacting with a pedestrian. Fig. \ref{fig:InteFea} visually shows the process of creating these interaction features for pedestrian $i$ in the middle for a sample scenario. In this figure pedestrians F and G, despite being close to pedestrian $i$, are not included in the interaction feature since they are not causing any collision risks.  

These interaction features will then be embedded according to equations \ref{eq:PedInt} and \ref{eq:VehInt} and used as an input to the LSTM.

\section{Experiment}

We have evaluated the effectiveness of our proposed method by first comparing the overall methods with a couple of baseline methods in the literature. Then we focus on the effectiveness of each of our introduced modules by conducting oblation studies.

\subsection{Dataset}

Since our method studies the effect of both pedestrian-pedestrian and pedestrian-vehicle interaction on the trajectory prediction of pedestrian agents, we require a dataset that contains both pedestrians and vehicles for our training and evaluation process. Meanwhile, we are willing to train our model in an environment where the interactions are less regulated by traffic rules and more governed through negotiation and social etiquette as these interaction patterns are more complicated to model compared to behaviours forced by traffic rules or emerged from strict environmental structures (e.g, crosswalks).
Therefore, datasets from shared space environments fit our objectives well since traffic rules and road markings are alleviated in these environments and all different types of agents are encouraged to share the same space~\cite{predhumeau2021pedestrian}. Here we use the HBS dataset \cite{pascucci2017discrete} where its real-world trajectory data has been captured from a shared space close to a busy train station in Germany from a bird's eye view. This dataset has already been used in other trajectory prediction methods in the literature~\cite{cheng2018modeling,cheng2020mcenet,johora2021transferability}. There are 338 vehicles, 1115 pedestrians, and 22 cyclists in the 30-minute video collected from this area and the trajectory data are available at 2 Hz. The dominant number of pedestrians in this dataset also suits our method which is focused on predicting only pedestrians' trajectories. Same as \cite{cheng2018modeling, pascucci2017discrete,cheng2018mixed}, we use the first 10 minutes (31\% of the dataset) as the test set and the remainder for training. Here we observe 6 time-steps covering 3 seconds of the trajectory and predict the next 6 time steps (3 seconds) of the trajectory, similar to~\cite{cheng2018modeling,cheng2018mixed}.

\subsection{Evaluation Metrics and Baselines} \label{Metrics}

In line with prior works we use the following metrics to evaluate the method and report the prediction errors.

\begin{itemize}
    \item Average displacement error (ADE): The Euclidean distance between predicted and ground truth positions at each time step averaged over all predicted time steps.
    \item Final Displacement error (FDE): The Euclidean distance between predicted and ground truth positions at the final predicted time step.
    \item Modified Hausdorff Distance (MHD): The largest distance between any two points on predicted and ground truth trajectories (without considering time step alignment).  
    \item Speed error (SE): The root mean square error (RMSE) between the speed of the predicted trajectory and the ground truth speed at each time step.
    \item Heading error (HE): The RMSE between the heading of the predicted trajectory and the ground truth heading at each time step.
\end{itemize}

We compare the performance of our method against the following baseline methods:

\begin{itemize}
    \item Linear Regression (LR): A linear regression model for each position dimension individually.
    \item Vanilla LSTM: A naive LSTM considering only trajectory history without the interaction effects from pedestrians and vehicles.
    \item Social LSTM: The LSTM model with the social pooling layer proposed by Alahi et al. \cite{alahi2016social} considering only pedestrian-pedestrian interaction.
\end{itemize}

The grid used in the Social LSTM model as the interaction encoding module is build based on relative position data. Thus, it is used as a baseline for evaluating our proposed interaction grid constructed using TTC information.

\subsection{Implementation Details}

The embedding layer for the spatial inputs and the two interaction feature all have a size of 64. We used a hidden state dimension of 128 for the LSTM module. The number of sectors ($n_{sector}$) for the polar collision grid was set to 8. A $TTC_{thre}$ of 9 seconds with a $d_{min}$ of 0.7 meters was used for the pedestrian-pedestrian interaction features. These values were 8 seconds and 1 meter for the pedestrian-vehicle interaction feature respectively. These values were selected after further tuning within a reasonable range according to the physical dimensions of the agents and the common personal space of humans \cite{gorrini2014experimental} as well as the agents’ minimum distance and speed ranges in the HBS dataset. Other hyperparameters of the model were also tuned by testing different values and selecting the one that resulted in the lowest error. A training batch size of 10 was used for 200 epochs. We used the RMS-prop optimizer with a learning rate of 0.001. The Nvidia GeForce RTX 2080 Ti GPU was used for the training and testing processes.  

\section{Results and Discussion}

\subsection{Quantitative Evaluation}

The performance of the trained models on the HBS test set are summarized in Table \ref{tab:OursVSbase}
Since the output of the deep-learning models is a distribution over the predicted position, for evaluation, same as \cite{zhang2022learning} we generated 20 samples for each model and reported the best result. Lower numbers are better. The best values among the different models for each metric are shown in boldface. The results show that our proposed method outperforms the baseline methods on all the evaluation metrics except for heading angle error. These results demonstrate the advantage of formulating interaction according to collision probability and approaching directions instead of only relying on relative distances. According to Table \ref{tab:OursVSbase}, Social LSTM has slightly higher prediction errors than the Vanilla LSTM which does not consider interactions. This is consistent with the result reported by others for these two models using other datasets \cite{gupta2018social,zhang2022learning}.

\begin{table}
\centering
\footnotesize
\caption{Prediction errors for baseline methods compared to the proposed method and its variants on the HBS dataset.} \label{tab:OursVSbase}
\resizebox{\columnwidth}{!}{\begin{tabular}{|l|c|c|c|c|l|}
\hline
\textbf{Models \textbackslash Metrics} & \textbf{ADE (m)} & \textbf{FDE (m)} & \textbf{MHD (m)} & \textbf{SE (m/s)} & \textbf{HE (\textdegree)} \\
\hline
Linear regression & 0.696 & 1.238 & 2.995 & 0.390 & 44.0\\
\hline
Vanilla LSTM & 0.305 & 0.676 & 2.855 & 0.240 & 32.7\\
\hline
Social LSTM & 0.309 & 0.677 & 2.852 & 0.244 & \textbf{31.5}\\
\hline
P-CollisionGrid & 0.304 & 0.664 & 2.811 & 0.235 & 32.6\\
\hline
V-CollisionGrid & 0.305 & 0.669 & 2.827 & \textbf{0.232} & 32.9\\
\hline
PV-CollisionGrid & \textbf{0.295} & \textbf{0.648} & \textbf{2.791} & 0.235 & 31.7\\
\hline
\end{tabular}}
\end{table}

\textbf{Oblation study:} To more clearly show the benefit of our proposed interaction modules for both pedestrian-pedestrian and pedestrian-vehicle interaction, we conducted an oblation study where we took out the interaction modules in a step-wise process and compared prediction errors with and without those modules in place.

Following this, we have trained our model by taking out the pedestrian-vehicle interaction feature and only keeping the collision grid associated with the pedestrian-pedestrian interaction. We called this the \textit{P-CollisionGrid}. We repeated the same process by this time taking out the pedestrian-pedestrian interaction feature and only keeping the pedestrian-vehicle collision grid and called the model \textit{V-CollisionGrid}. The prediction errors of these two models along with our complete model \textit{PV-CollisionGrid} are also reported in Tabel \ref{tab:OursVSbase}. 

The results show that adding each of the interaction modules positively improves the prediction accuracy of the model in terms of ADE, FDE, MHD and heading error. This indicates the effectiveness of our interaction module in capturing the influence of potentially conflicting agents on the predicted trajectory of a pedestrian. However, the lowest speed error is for the model that only considers pedestrian-vehicle interaction. This could be due to the high influence that vehicles have on pedestrians' speed adjustment along their trajectory especially when a pedestrian stops for giving way to a vehicle.

\textbf{Interacting agent selection:} For more specifically analyzing the effect of our proposed method for the selection of interacting agents based on TTC, we compared two versions of the Social LSTM baseline model. The first version is the original one where the interaction feature is constructed using agents closer than a specified distance. In the second version, the neighbouring agents are first filtered based on our proposed TTC criteria and then the original social pooling module \cite{alahi2016social} is constructed considering those agents with collision risks.

The performance evaluation of these two versions in Table \ref{tab:Selection} shows that filtering the agent to consider only the ones in a conflict course with the target agent will actually improve the prediction accuracy. According to this result, removing the information of those agents that do not produce a collision risk on the target agent seems to be not only harmless but also beneficial as the target agent's trajectory will not be affected by those agents. Therefore, our proposed method of selecting interacting agents based on collision risk seems to outperform the distance-based methods by concentrating on those agents that actually make the target agent adjust its path for avoiding possible collisions.

\begin{table}
\centering
\footnotesize
\caption{The effect of our proposed interacting agent selection based on collision checks (used as a filter) on the performance of the Social LSTM model reported on HBS dataset. 
} \label{tab:Selection}
\resizebox{\columnwidth}{!}{\begin{tabular}{|l|c|c|c|c|l|}
\hline
\textbf{Models \textbackslash Metrics} & \textbf{ADE (m)} & \textbf{FDE (m)} & \textbf{MHD (m)} & \textbf{SE (m/s)} & \textbf{HE (\textdegree)} \\
\hline
Social LSTM & 0.309 & 0.677 & 2.852 & 0.244 & \textbf{31.5}\\
\hline
\begin{tabular}{@{}l@{}}Social LSTM \\ +Filtered interaction \end{tabular}  & \textbf{0.298} & \textbf{0.658} & \textbf{2.827} & \textbf{0.234} & 32.3\\
\hline
\end{tabular}}
\end{table}

\begin{figure}[b!]
  \centering
  \includegraphics[width=0.8\linewidth]{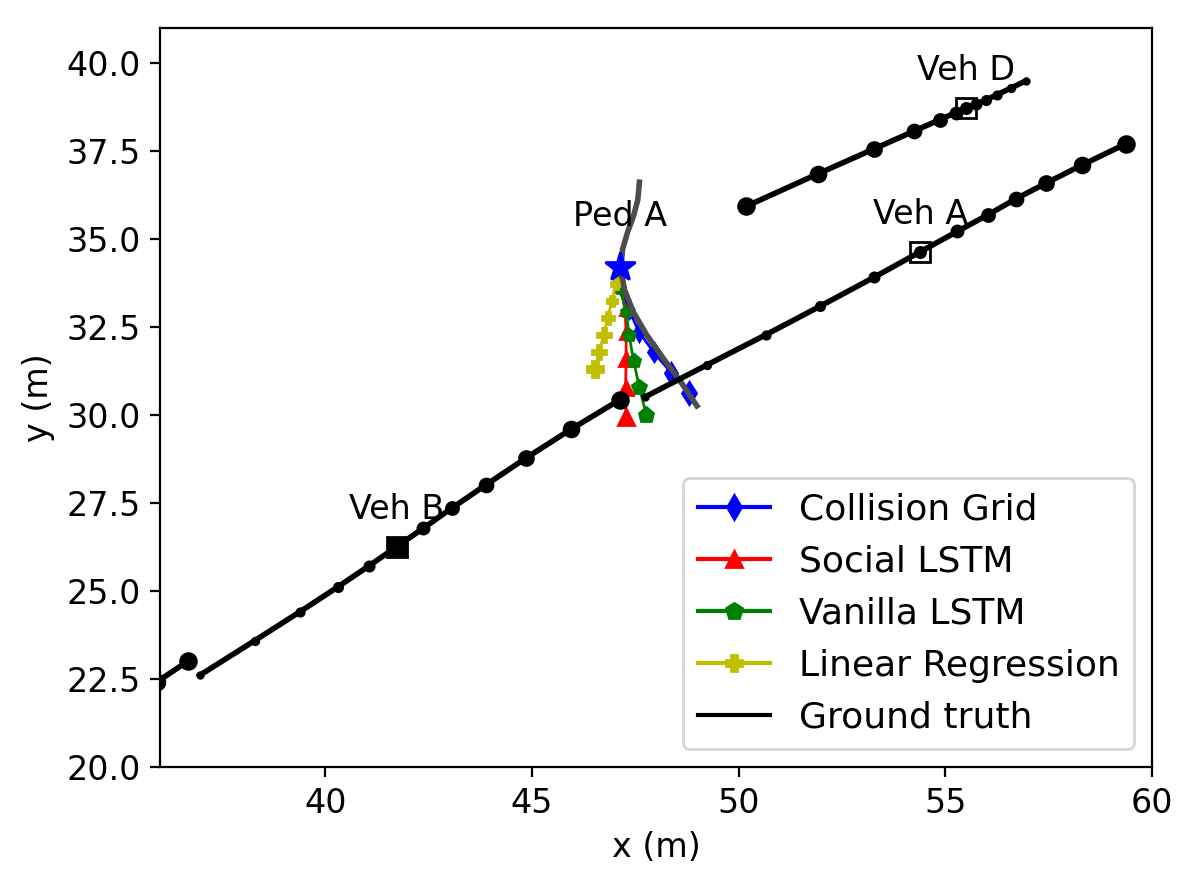}
  \caption{The predicted trajectories for our method compared to the baseline methods for a pedestrian-vehicle interaction scenario in the HBS test dataset. A marker with a larger size represents a later time step along the trajectory. The position of the vehicles at the last time step of the observation period is indicated with a square marker. Our method detects the target pedestrian A (indicated with start) to have interaction with vehicle B at this time step according to the TTC criteria. This identified interacting vehicle is indicated with a filled square marker at this time step. Considering this interacting vehicle our method has predicted a trajectory that avoids the predicted collision while being close to the ground truth trajectory.}
  \label{fig:P-V}
\end{figure}

\begin{figure*}[b!]
\centering
\subfigure[]{
\includegraphics[height=4.0cm]{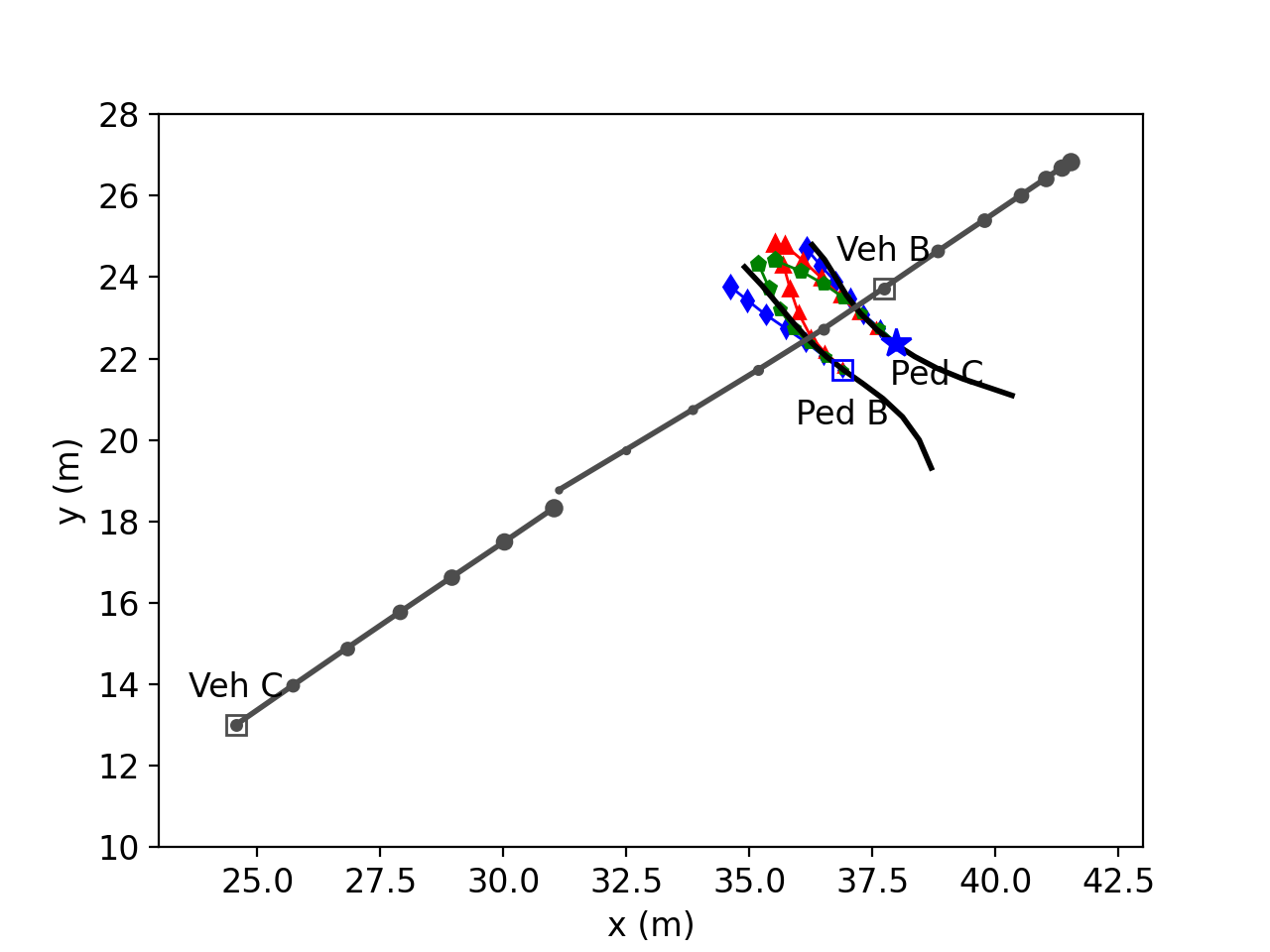}
\label{fig:1}
}
\subfigure[]{
\includegraphics[height=3.6cm]{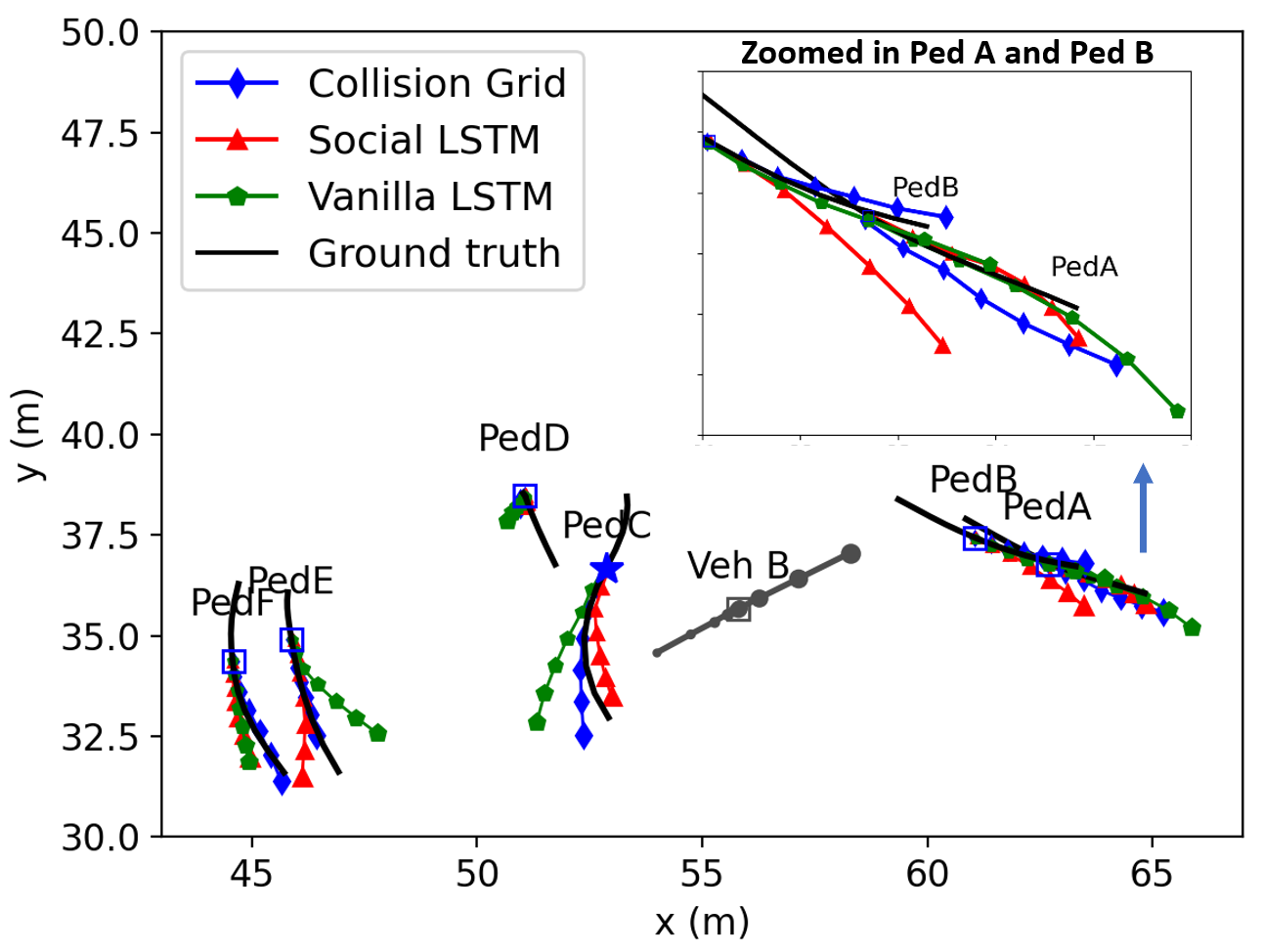}
\label{fig:2}
}
\subfigure[]{
\includegraphics[height=4.0cm]{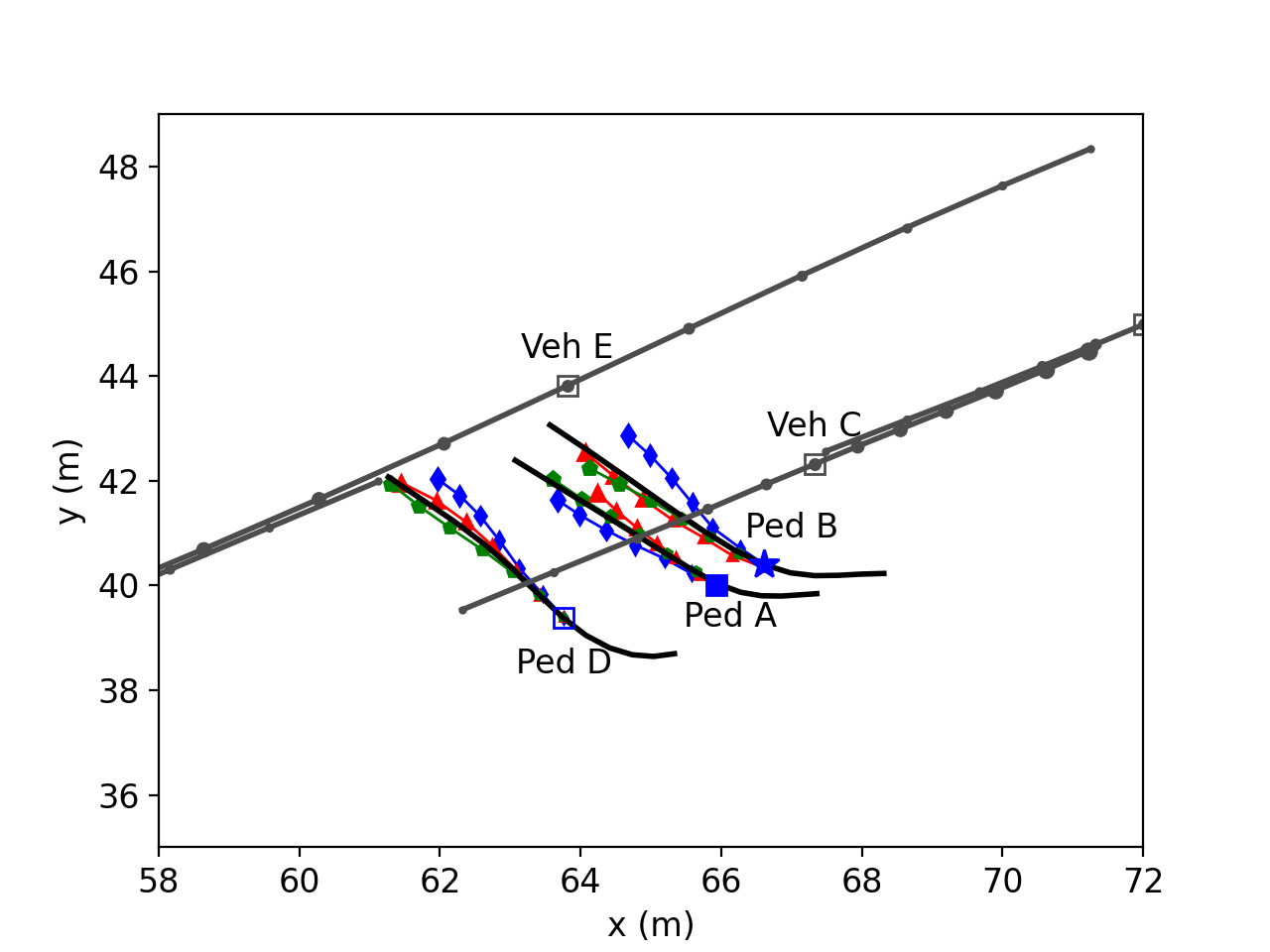}
\label{fig:3}
}
\caption{The predicted trajectories for our method compared to the deep-learning-based baseline methods in the HBS dataset. Our predictions overall are closer to the ground truth compared to the baselines. Case (c) is an example with a non-accurate prediction from our \textit{Collision Grid} method where a probable group interaction between Ped A and Ped B is mistakenly detected as a possible collision ending up with a prediction for Ped B that  deviates to the right of the ground truth data. Interacting agents with a sample target pedestrian (shown with star) are shown with a filled square marker.
}
\label{fig:more}
\end{figure*}

\begin{figure}[t!]
\centering
\subfigure[]{
\includegraphics[height=3.9cm]{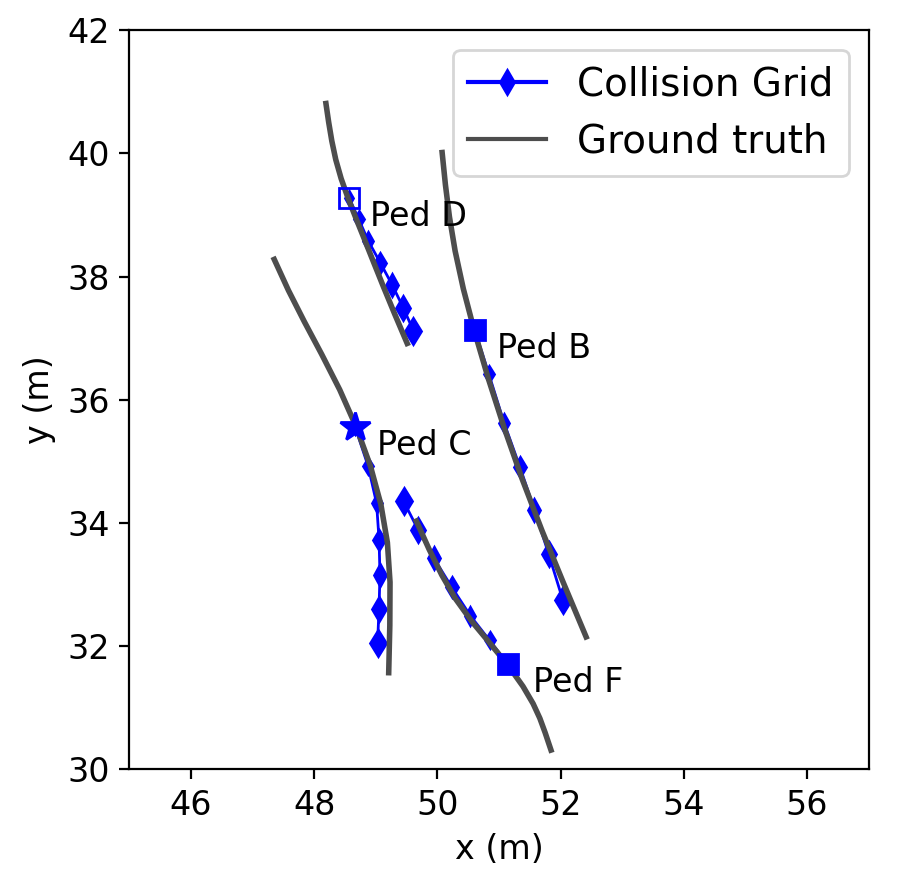}
\label{fig:423CG}
} 
\subfigure[]{
\includegraphics[height=3.9cm]{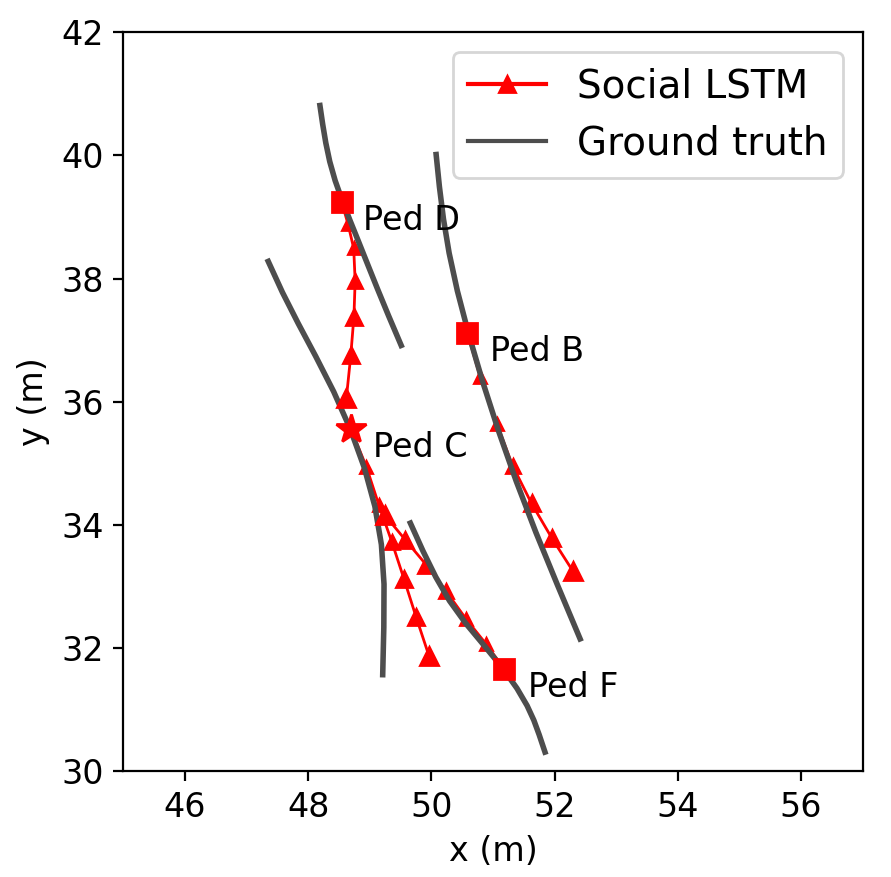}
\label{fig:423SL}
} 
\caption{Prediction of our Collision Grid method compared to Social LSTM for a pedestrian-pedestrian interaction scenario. Our method produces a prediction closer to the ground truth for the pedestrians such as Ped C. This is done by detecting Ped C's interacting agents (Ped F and Ped B) more accurately while relying on time to collision for interaction modelling as opposed to the Social LSTM model that considers all agents closer to the target pedestrian than a threshold distance regardless of their moving direction which happens to be all the agents present in the scene here. Interacting agents for Ped C (target) are indicated with a filled square marker at time step $t_{obs}$ while markers for non-interacting agents are unfilled.}
\label{fig:P-P}
\end{figure}

\subsection{Qualitative Evaluation}

To qualitatively analyze the predicted trajectory of our model compared to other baselines, we plotted the prediction outputs for a couple of interactive scenarios in the test set as shown in Fig. \ref{fig:P-V}-\ref{fig:P-P}.

In the scenario shown in Fig. \ref{fig:P-V}, our method detects vehicle B as an interacting agent with pedestrian A in the last observed timestep ($t_{obs}$), meaning that a collision will take place if the two agents continue their velocity as it is at time $t_{obs}$.
Having identified this interaction, our method then considers Vehicle B's information on TTC and approach angle when predicting pedestrian A's future trajectory. Therefore, as shown in Fig. \ref{fig:P-V}, our method predicts an avoidance manoeuvre which is close to the ground truth. This is while the Social SLTM method which does not consider the pedestrian-vehicle interaction, predicts a trajectory that ends up in a collision with the vehicle at the last time step of the prediction window. This supports the effectiveness of our pedestrian-vehicle interaction module. 

Fig. \ref{fig:P-P} shows another example focused on pedestrian-pedestrian interactions and compares the prediction of our Collision Grid method with the predictions of the Social LSTM method that also models the interaction between pedestrians. Focusing on Pedestrian C as a target agent which is indicated with a star marker at time step $t_{obs}$, we show all its neighbours at the same time step with a square marker. In each of the methods, the neighbouring agents that will be considered for constructing the interaction feature are indicated with a filled square marker. According to Fig. \ref{fig:P-P} while the Social LSTM considers all the agents present in this scene in building the interaction, our method only detects pedestrian F and B as interacting agents for pedestrian C and therefore relies only on these pedestrians' information for constructing the interaction feature. This results in a more accurate prediction for Pedestrian C's future trajectory with less deviation from the ground truth compared to the Social LSTM method. This shows the effectiveness of our proposed interacting agent selection process that focuses specifically on neighbouring agents that can actually affect the target agent's trajectory due to being in a potential collision course, as opposed to other methods such as Social LSTM that consider all agents that are closer than a distance to the target agent regardless of their moving direction. This latter approach can introduce extra not useful information for interaction consideration that can lead to an undesired prediction as seen in Fig. \ref{fig:P-P} for Social LSTM.  

More qualitative examples of our method's prediction compared to other baseline methods can be found in Fig. \ref{fig:more} for other scenarios.
There are cases where our method has more deviation from the ground truth compared to other methods. Fig. \ref{fig:3} provides an example in which two pedestrians (Ped A and Ped B) are walking together as a group. For this case our method has a false positive collision detection and predicts a trajectory for Ped B that is slightly to the right of the ground truth, getting away from Ped A, instead of predicting close parallel paths for these two pedestrians. However, despite the poor prediction in these particular cases, our algorithm produces less error on average compared to other baselines' predictions as seen in Table \ref{tab:OursVSbase}. In future, the prediction of our algorithm can be further improved by adding a group detection for filtering out possible false collision detection in the model.

\section{Limitation and future work}

Our work had several limitations. The goal was to model the interaction effect between pedestrians and vehicles to predict future trajectories of pedestrians. These complicated interactions mostly happen in environments that do not segregate the operational space of the pedestrians and vehicles, e.g., shared spaces. 
Therefore, we limited our datasets to shared space environments that greatly embed these high-interactive behaviours. However, only a few datasets exist from these shared spaces while in some of them the number of data points and trajectories are limited. Thus, we only trained and tested our algorithm on one dataset (HBs) which outperformed the baselines. Future work can benefit from testing on other datasets to confirm whether the results can be generalized to other environments. Also, for focusing on patterns that could be derived from only trajectory data, we did not consider the effect of the environment's structure. Future work can take the map of the environment as another input to study the effects of these other factors.

\section{Conclusion}

We proposed the use of collision-risk information for encoding the interaction effects for pedestrian trajectory prediction in an environment shared with vehicles. Interacting pedestrians and vehicles with a target pedestrian were first selected based on the heuristic of Time-to-Collision (TTC). A novel polar collision grid map for each target pedestrian then encoded the interaction effects using the information of TTC and the approach direction of these interacting agents.

Our results showed that by filtering the neighbouring agents for focusing only on those agents that could actually affect the target pedestrians through posing a collision risk, we may better capture the interaction effects on the predicted trajectories. Also, by directly using the information on the approaching risk direction and the criticality of the situation using TTC, we were able to predict the corresponding changes in the trajectory more accurately than the baseline methods. According to the results, we believe that providing the prediction network with more relevant engineered features can guide deep-learning-based prediction methods to better learn the patterns underlying  the trajectories, leading to predictions that are closer to the ground truth.






\bibliographystyle{IEEEtran}
\bibliography{IEEEabrv,mybibfile}

\end{document}